\documentclass[10pt,twocolumn,letterpaper]{article}

\usepackage[pagenumbers]{cvpr}
\definecolor{cvprblue}{rgb}{0.21,0.49,0.74}
\usepackage[pagebackref,breaklinks,colorlinks,allcolors=cvprblue]{hyperref}

\usepackage{multirow}

\def\confName{CVPR}
\def\confYear{2027}

\title{TSFFormer: A Frequency-Domain Spike Transformer for Green Human Activity Recognition}

\author{Jiale Deng\\
Xidian University\\
Xi’an 710071, China\\
{\tt\small 25011211021@stu.xidian.edu.cn}
\and
Shuiying Xiang\\
Xidian University\\
Xi’an 710071, China\\
{\tt\small syxiang@xidian.edu.cn}
}

\begin{document}
\maketitle
\begin{abstract}

Wearable Inertial Measurement Units (IMUs) enable pervasive Human Activity Recognition (HAR), but traditional Deep Neural Networks (DNNs) suffer from an ``arithmetic-latency deadlock'' due to dense floating-point MAC operations. While Spiking Neural Networks (SNNs) offer ultra-low power, existing ones face spatial blind spots, rigid neurons, and gradient degradation. To address these, we propose \textbf{TSFFormer}, a purely spiking Fourier Transformer for Green HAR. It introduces two key innovations: \textbf{(1)} a \textbf{MultiHeadFourier} attention that performs frequency-domain global mixing, reducing complexity from \(O(V^2)\) to \(O(V\log V)\), eliminating spatial blind spots, and saving \(\sim\)50\% GPU memory; and \textbf{(2)} a dual-compartment \textbf{Temporal Spiking LIF (TSLIFNode)} that utilizes dynamic thresholds to adapt to non-stationary rhythms and mitigate gradient degradation. Extensive experiments on 8 public IMU benchmarks demonstrate that TSFFormer surpasses state-of-the-art SNNs and even the classic DeepConvLSTM, achieving a \(\sim\)3\(\times\) faster training speed with an energy footprint of 0.1 pJ. Ablation studies confirm the synergistic effect of our modules, providing a highly accurate, ultra-low-power event-driven paradigm for wearable sensing.

\end{abstract}    
\section{Introduction}

HAR using IMUs is a cornerstone of pervasive computing\cite{zheng2026towards}. Over the past decade, the field has evolved from hand-crafted features and shallow classifiers to DNNs and, more recently, heavy Transformer-based foundation models \cite{zhang2024unimts}. However, this relentless pursuit of accuracy has spawned a critical bottleneck in edge deployment: the \textit{arithmetic-latency deadlock}. DNNs rely on dense 32-bit floating-point (FP32) Multiply-Accumulate (MAC) operations (\(\sim\)4.6 pJ per step) and rigid ``buffer-then-compute'' paradigms, resulting in prohibitive power consumption and observation latency that make always-on HAR impractical.

SNNs offer a neuromorphic alternative by replacing dense MACs with sparse Accumulate (AC) operations (\(\sim\)0.1 pJ) \cite{bellec2020solution}. Despite this efficiency, applying SNNs to IMU-based HAR faces severe limitations: (1) \textit{Spatial blind spots}: naive SNNs flatten sensor data, ignoring human biomechanical topology\cite{zheng2026towards}. (2) \textit{Neural rigidity}: standard Leaky Integrate-and-Fire (LIF) neurons use fixed thresholds, failing to adapt to non-stationary motion rhythms\cite{feng2025ts}. (3) \textit{Gradient degradation}: training long sequences via BPTT forces models to rely on long integration windows, precluding low-latency inference.

To bridge these gaps, we propose \textbf{TSFFormer}, a purely spiking Fourier Transformer for Green HAR. It integrates a \textbf{Temporal Spiking LIF (TSLIF)} neuron to dynamically modulate firing thresholds, and a \textbf{MultiHeadFourier} attention \cite{gan2026caracal} to efficiently capture global spatio-temporal dependencies in the frequency domain. Combined with a non-downsampling Patch Embedding (MS-SPS) to preserve spatial topology, TSFFormer fundamentally resolves the aforementioned limitations. Our main contributions are:

\begin{itemize}

    \item \textbf{Purely Spiking Fourier Paradigm}: We propose TSFFormer, a HAR architecture that deeply fuses spiking neural networks with Fourier attention. This model significantly reduces the operation energy consumption and greatly improves the inference speed.
    
    \item \textbf{Dual-Adaptive Spatio-Temporal Core}:We used a dual adaptive core. The TSLIF neuron prevents catastrophic accuracy drops caused by fixed thresholds. Meanwhile, MultiHeadFourier reduces complexity from \(O(V^2)\) to \(O(V\log V)\), accelerating training by \(\sim\)3\(\times\) and saving \(\sim\)50\% GPU memory.
    
    \item \textbf{Topology-Preserving Sparse Computation}: We adopt MS-SPS for channel mapping and keep the vertex dimension unchanged, so as to completely preserve the spatial structure of sensor nodes. Combined with the sparse event-driven characteristics of spiking MLP, the model greatly reduces redundant computations while maintaining high accuracy.
\end{itemize}
\section{Related Work}

\subsection{Evolution of IMU-based HAR Architectures}
IMU-based  HAR has evolved from early classifiers with hand-crafted features to automated representation learning. Deep architectures, notably CNNs and recurrent networks (e.g., DeepConvLSTM\cite{singh2020deep}), established strong baselines by fusing local spatial extraction with temporal modeling. Subsequent research integrated attention mechanisms and advanced macro-architectures \cite{ronald2021isplinception} to capture salient spatio-temporal segments. Recently, the field has embraced large-scale foundation models and LLM-driven semantic reasoning \cite{zhang2024unimts}  to handle extreme cross-domain variability. Among these, LIMU-BERT stands out as a representative IMU-specific foundation model that applies self-supervised learning to unlabeled sensor data for generalized representation\cite{xu2021limu}.

\subsection{Spiking Neural Networks and Energy-Efficient Computing}
To circumvent the energy bottleneck, SNNs have emerged as a promising ultra-low-power paradigm. By processing asynchronous binary events, SNNs\cite{fang2023spikingjelly} replace power-hungry MACs with efficient AC operations. Driven by Surrogate Gradient (SG) methods and Spatio-Temporal Backpropagation (STBP), deep SNNs have achieved notable progress \cite{bellec2020solution}. The fundamental computational unit is typically the LIF neuron, whose discrete-time dynamics are\cite{nicola2015leukemia}:
\begin{equation}
U[t] = \tau U[t-1](1 - S[t-1]) + WX[t]
\end{equation}
\begin{equation}
S[t] = H(U[t] - V_{th})
\end{equation}
where \(U[t]\) is the membrane potential, \(\tau\) is the decay factor, \(X[t]\) is the input current, \(W\) represents synaptic weights, and \(S[t]\) is the binary output spike triggered when \(U[t]\) exceeds the firing threshold \(V_{th}\). \(H(\cdot)\) denotes the Heaviside step function. 

Despite its success, deploying SNNs for IMU-based HAR exposes three critical gaps: (1) \textbf{Spatial blind spots}: existing SNNs typically flatten multi-sensor series into independent 1D channels, lacking structural capability to explicitly route physical topologies. (2) \textbf{Neural rigidity}: standard LIF neurons rely on a fixed \(V_{th}\). Adaptive variants (e.g., ALIF\cite{bellec2018long}) introduce heuristic threshold decay but lack causal modulation for rapidly changing human motion rhythms\cite{feng2025ts}. (3) \textbf{Gradient degradation}: optimizing deep SNNs over long sequences via BPTT forces networks to rely on long integration windows, precluding ultra-low-latency prediction.

\subsection{Spiking Transformers and Frequency-Domain Attention}
Recent Spiking Transformers\cite{zhou2022spikformer, yao2023spike, zhou2024qkformer,zhou2023spikformer} introduce attention-style mixing into SNNs for global modeling. However, they mostly focus on static images, relying on dot products or linear projections that struggle to capture global dependencies in the frequency domain, often losing biomechanical correlations due to flattened inputs. Conversely, frequency-domain modeling in time-series analysis offers unique advantages. The Fourier transform converts temporal convolutions into frequency-domain multiplications\cite{lee2021fnet,lee2025spiking}, capturing long-range dependencies with lower computational complexity. Combining multi-head mechanisms with Fourier transforms (MultiHeadFourier) avoids explicit sequence-length squared complexity while retaining sensitivity to periodic patterns\cite{gan2026caracal}. However, seamlessly integrating frequency-domain attention into SNNs while maintaining event-driven sparse computation remains an open problem.

\subsection{Adaptive Spiking Neurons and Temporal Dynamics}
The fixed threshold of standard LIF neurons struggles to adapt to non-stationary motion rhythms. While variants like ALIF adjust sensitivity via threshold decay, they rely on heuristic rules and lack explicit modeling of causal temporal contexts. TSLIF neurons introduce time-dependent threshold modulation, simulating the dynamic threshold characteristics of biological neurons to enable precise feature extraction across different motion stages\cite{2024Temporal}. However, effectively combining TSLIF with Spiking Transformer architectures while maintaining gradient stability during training requires further exploration\cite{capone2023beyond}.

\section{Methodology}

\subsection{Overall Architecture}
To achieve high-accuracy, low-power HAR in wearable inertial sensing, we propose TSFFormer---a Green HAR architecture that fuses SNNs, TSLIF neurons, and MultiHeadFourier. Formally, given a raw multi-node IMU tensor \(X \in \mathbb{R}^{T \times B \times C \times V}\) (time steps \(T\), batch size \(B\), sensor channels \(C\), physical nodes \(V\)), the model processes data through a purely spike-driven forward pipeline and outputs classification predictions along the temporal dimension.

\begin{figure*}[t]
    \centering
    \includegraphics[width=\textwidth]{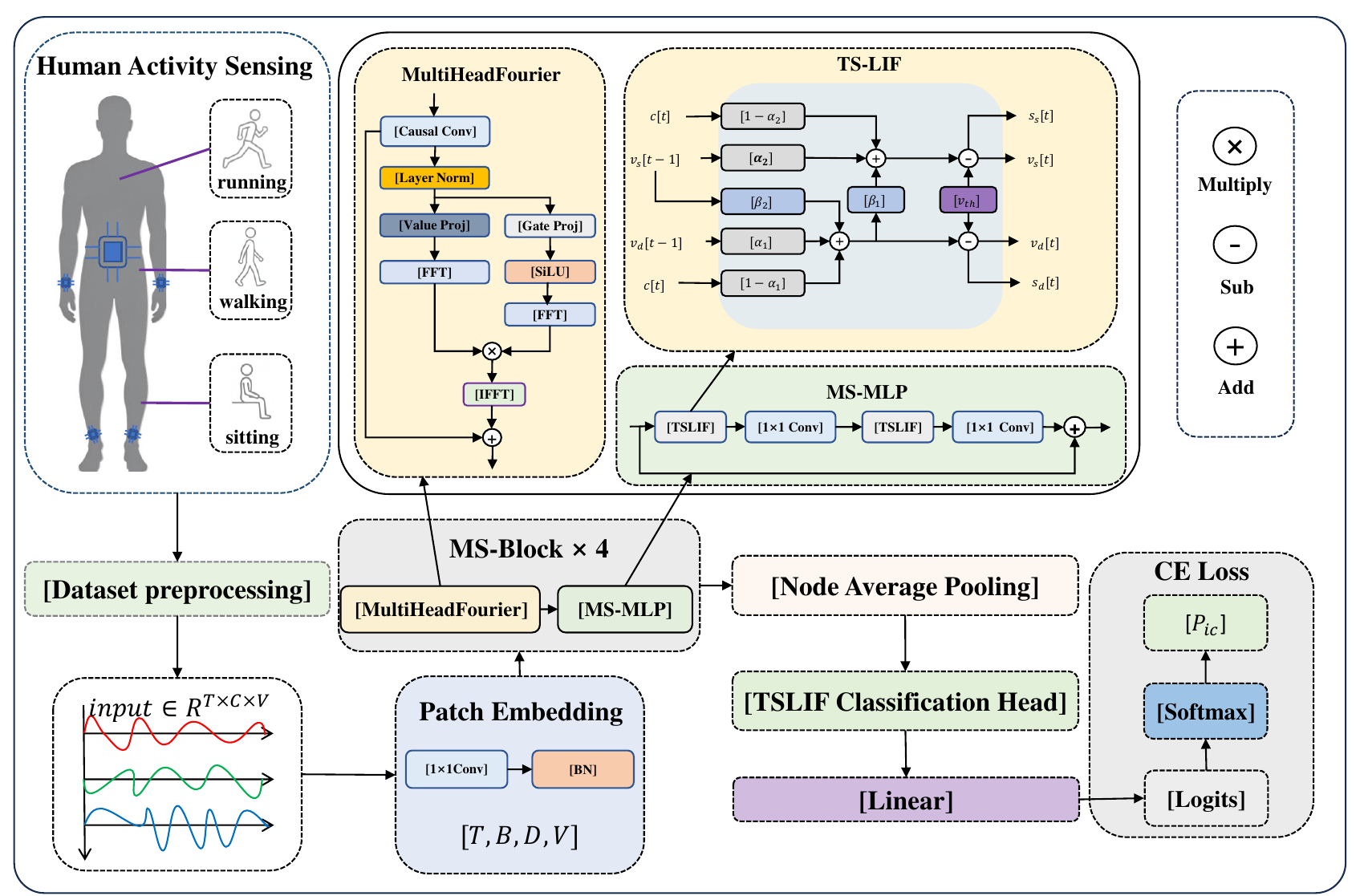} 
    \caption{The overall structure of our proposed TSFFormer framework.}
    \label{fig:arch}
\end{figure*}

As shown in Figure \ref{fig:arch}, TSFFormer consists of several parts: the perception front-end and spiking embedding map continuous inputs into high-dimensional spike sequences; spatio-temporal spiking Transformer blocks are stacked for \(L\) layers, each containing MultiHeadFourier attention and a spiking MLP; temporal dimension readout aggregates features along the vertex dimension; and the classification head averages predictions over the time dimension. The entire network replaces dense floating-point MACs with event-driven sparse ACs, utilizes the dynamic threshold modulation of TSLIF neurons to adapt to non-stationary motion rhythms, and efficiently captures global joint correlations through frequency-domain attention.

\subsection{Perception Front-end and Spiking Embedding}
The raw IMU tensor \(X \in \mathbb{R}^{T \times B \times C \times V}\) is first mapped to the embedding space via a Simple MS-SPS.
MS-SPS employs a single-layer 1D convolution (kernel size 1) and Batch Normalization (BatchNorm1d) to achieve channel transformation:
\begin{equation}
    X_{emb} = \mathrm{BN}(\mathrm{Conv1d}_{1\times 1}(X)) \in \mathbb{R}^{T \times B \times D \times V}
\end{equation}
where \(D\) is the embedding dimension. This operation does not perform spatial downsampling, fully preserving the biomechanical topology of the vertex dimension. For the first layer, no spiking activation is applied; subsequent layers pass through a TSLIF neuron before convolution to introduce spiking sparsity. The output \(X_{emb}\) serves as the input to the first Transformer block.

\subsection{Spatio-Temporal Spiking Transformer Block}
The network backbone consists of \(L\) stacked MS-Blocks. Each block contains two sub-layers: MultiHeadFourier attention and a spiking MLP, both using residual connections and DropPath regularization. The input tensor maintains the shape \(T \times B \times D \times V\), with all operations performed independently along the temporal dimension.

\subsubsection{Multi-Head Fourier Attention}
To efficiently capture global dependencies between vertices and avoid the quadratic complexity of traditional dot-product attention, we adopt Multi-Head Fourier attention. Given input \(X \in \mathbb{R}^{T \times B \times D \times V}\), we first reshape it into \(B \times T\) independent sequences:
\begin{equation}
\resizebox{0.90\linewidth}{!}{$
    X_{seq} = \mathrm{reshape}(X.\mathrm{permute}(1,0,3,2)) \in \mathbb{R}^{(BT) \times V \times D}$}
\end{equation}
meaning each time slice and sample is processed independently, with the vertex dimension \(V\) as the sequence length and \(D\) as the feature dimension. Then, MultiHeadFourier executes the following steps:

\begin{itemize}
    \item \textbf{Local causal convolution:} Use a depthwise separable 1D convolution (kernel size 3, groups \(D\)) to extract local temporal inductive bias, with left padding to ensure causality:
    \begin{equation}
        X_{conv} = \mathrm{Conv1d}_{\mathrm{depthwise}}(X_{seq})
    \end{equation}
    \item \textbf{Layer Normalization:} Apply LayerNorm to the convolution output.
    \item \textbf{Content and Gating Stream Projections:}
    \begin{itemize}
        \item Content stream:\\ \(V = W_V(\mathrm{LN}(X_{conv})) \in \mathbb{R}^{(BT) \times V \times D}\), split into \(H\) heads with dimension \(d_h = D/H\).
        \item Gating stream:\\ \(G = W_{G2}(\mathrm{SiLU}(W_{G1}(\mathrm{LN}(X_{conv}))))\), where \(W_{G2}\) is a grouped 1D convolution (groups \(H\)), also split into heads.
    \end{itemize}
    \item \textbf{Frequency-Domain Causal Mixing:} Apply real Fast Fourier Transform (rFFT) to both content and gating streams, padding length to \(N=2V\):
    \begin{equation}
        \mathcal{V} = \mathrm{rFFT}(V, n=N), \quad \mathcal{G} = \mathrm{rFFT}(G, n=N)
    \end{equation}
    Multiply element-wise in the frequency domain:
    \begin{equation}
        \mathcal{X} = \mathcal{V} \odot \mathcal{G}
    \end{equation}
    Then inverse transform back to the time domain and truncate to the original length \(V\):
    \begin{equation}
        X_{mixed} = \mathrm{irFFT}(\mathcal{X}, n=N)[:,:,:V,:]
    \end{equation}
    \item \textbf{Output Projection:} Concatenate the multi-head results and pass through a linear layer:
    \begin{equation}
        Y = W_O(\mathrm{reshape}(X_{mixed})) \in \mathbb{R}^{(BT) \times V \times D}
    \end{equation}
\end{itemize}

Finally, restore the shape to \(T \times B \times D \times V\). This design leverages the property that frequency-domain multiplication is equivalent to time-domain convolution, achieving a global receptive field with \(O(V\log V)\) complexity\cite{gan2026caracal}, while adaptively adjusting frequency-domain responses via the gating mechanism to enhance modeling of periodic motion patterns.

\subsubsection{Spiking MLP}
The spiking MLP is responsible for channel mixing and non-linear transformation. The input \(X \in \mathbb{R}^{T \times B \times D \times V}\) passes through two layers sequentially:
\begin{align}
    X_1 &= \mathrm{TSLIF}(\mathrm{BN}_1(\mathrm{Conv1d}_{1\times1}(X))) \\
    X_2 &= \mathrm{BN}_2(\mathrm{Conv1d}_{1\times1}(X_1)) \\
    Y &= \mathrm{TSLIF}(X_2)
\end{align}
Before each convolution, the \(T\) and \(B\) dimensions are flattened to adapt to 1D convolution. The hidden dimension is \(D_{hid} = \lfloor D \cdot \mathrm{mlp\_ratio} \rfloor\). Both TSLIF neurons act on the channel dimension, and the output maintains \(T \times B \times D \times V\).

\subsubsection{TSLIFNode}
TSLIFNode is a dual-compartment spiking neuron whose dynamics simulate the dendrite-soma coupling and dynamic threshold modulation of biological neurons\cite{feng2025ts}.
Let the input at time step \(t\) be \(c_t\), and the membrane potential states be \(v_d\) (dendritic compartment) and \(v_s\) (somatic compartment). The update rules are as follows:
\begin{equation}
\resizebox{0.88\linewidth}{!}{$ v_d[t] = \alpha_1 v_d[t-1] + \beta_1 v_s[t-1] + (1-\alpha_1)c_t - \gamma_1 s_d[t-1]$}
\end{equation}
\begin{equation}
\resizebox{0.88\linewidth}{!}{$ v_s[t] = \alpha_2 v_s[t-1] + \beta_2 v_d[t] + (1-\alpha_2)c_t - \gamma_2 s_s[t-1] ]$}
\end{equation}
\begin{equation}
s_d[t] = H(v_d[t] - v_{th}), \quad s_s[t] = H(v_s[t] - v_{th})
\end{equation}
where \(\alpha_1, \alpha_2, \beta_1, \beta_2\) are learnable decay factors, \(\gamma_1, \gamma_2\) are inter-compartment coupling strengths, and \(v_{th}\) is the firing threshold:
\begin{equation}
    v_{th} = V_{th} \cdot (2 - g_t)
\end{equation}

\begin{figure}[t]
    \centering
    \includegraphics[width=\linewidth]{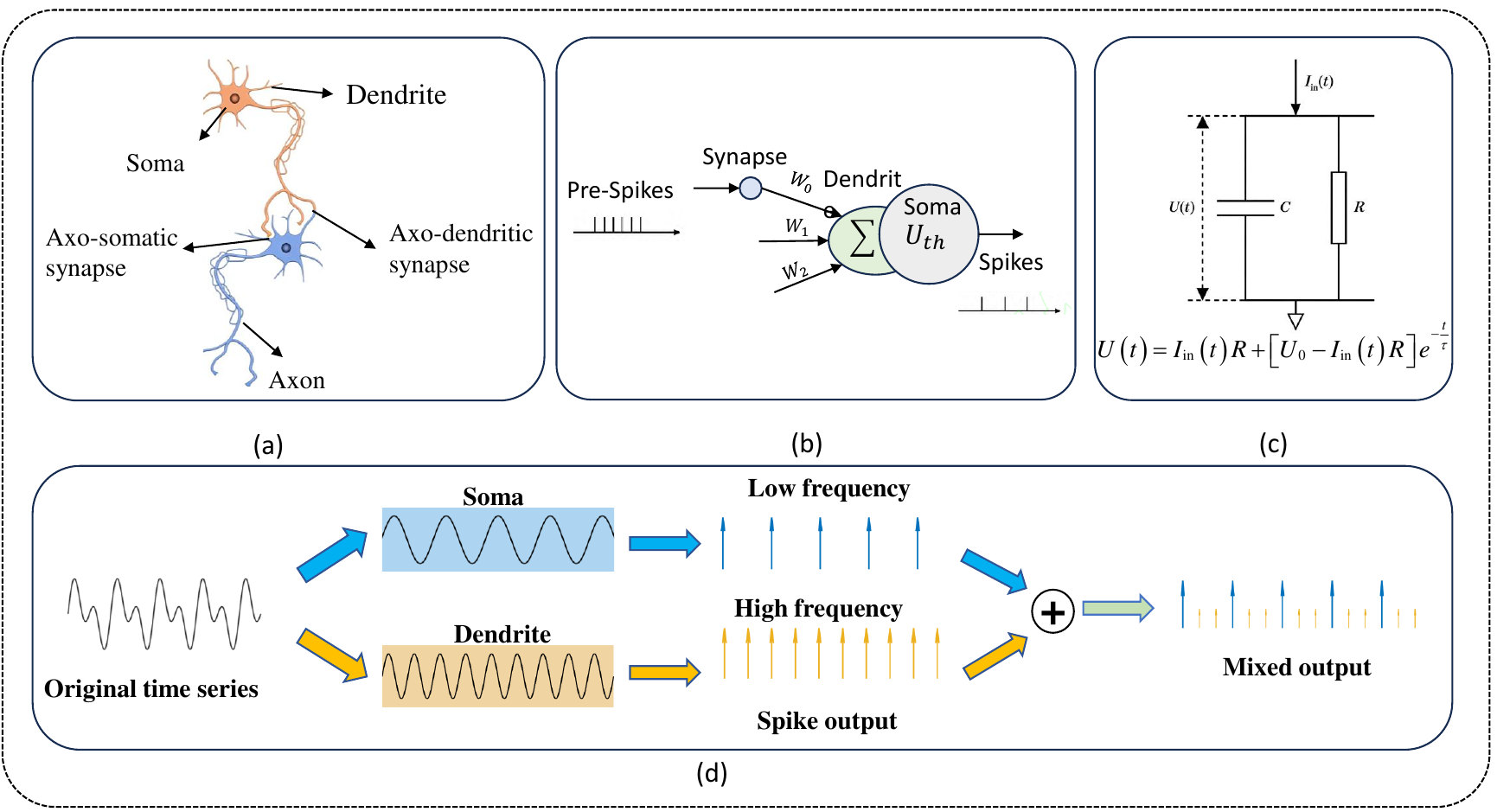} 
    \caption{(a) Structural organization of neuronal signal transmission: focusing on axon-soma synapses and axon-dendrite synapses; (b) SNN neuron structure; (c) Circuit diagram and formula of the LIF model; (d) Time-series decomposition and spike generation in the TS-LIF model.}
    \label{fig:tslif}
\end{figure}

where \(V_{th}\) is the baseline threshold, and \(g_t\) is an optional external gating signal (used for contextual modulation in this paper). \(s_d[t]\) and \(s_s[t]\) are the spike outputs of the dendritic and somatic compartments, respectively, and \(H(\cdot)\) is the Heaviside step function, with ATan surrogate gradients used in actual training. This design enables neurons to dynamically adjust firing sensitivity based on historical activity, lowering the threshold during explosive movements to enhance response and raising it at rest to suppress noise, thereby adapting to non-stationary motion rhythms.

\subsection{Temporal Dimension Readout and Classification Head}
After \(L\) MS-Blocks, the feature tensor \(X \in \mathbb{R}^{T \times B \times D \times V}\) is first averaged over the vertex dimension to aggregate information from all physical nodes:
\begin{equation}
    X_{pool} = \frac{1}{V}\sum_{v=1}^{V}X[:,:,:,v] \in \mathbb{R}^{T \times B \times D}
\end{equation}
Then, it passes through a TSLIF neuron to generate spike activations, followed by a linear classification head mapping to the class space:
\begin{equation}
    \hat{Y} = W_{cls} \cdot \mathrm{TSLIF}(X_{pool}) \in \mathbb{R}^{T \times B \times N_c}
\end{equation}
Finally, averaging over the temporal dimension yields the classification prediction for each sample:
\begin{equation}
    \hat{y} = \frac{1}{T}\sum_{t=1}^{T}\hat{Y}[t] \in \mathbb{R}^{B \times N_c}
\end{equation}
This readout strategy aggregates predictions over time, avoiding latency issues associated with relying only on the final time step, while further reducing computation via the sparsity of spiking neurons.

\section{Experiments}
To comprehensively evaluate TSFFormer, we conduct extensive experiments on eight public IMU benchmarks. Our evaluation is guided by the following research questions (RQs):

\begin{itemize}
    \item \textbf{RQ1 (Overall Performance):} How does TSFFormer compare against state-of-the-art DNNs and SNNs across various scenarios?
    \item \textbf{RQ2 (Training Efficiency):} Can TSLIF neurons and Multi-Head Fourier attention effectively accelerate convergence and mitigate gradient degradation?
    \item \textbf{RQ3 (Component Contribution):} How much do the key designs (e.g., spiking, TSLIF, frequency-domain mixing) specifically contribute to the final performance?
\end{itemize}

\subsection{Experimental Setup}
\subsubsection{Datasets}
To ensure robust evaluation across different wearable scenarios, we selected 8 public IMU datasets covering multi-node, single-node, clinical, and elderly populations. Table~\ref{tab:dataset} summarizes the detailed configurations of these datasets.

\begin{itemize}
    \item \textbf{Multi-node sensor environments}: PAMAP2 \cite{reiss2012introducing}, TNDA-HAR, and HuGaDB \cite{chereshnev2017hugadb} contain multiple distributed sensor nodes, with TNDA-HAR comprising 50 subjects and 6 asymmetric sensor nodes, making it the largest multi-node benchmark.
    \item \textbf{Complex human activities}: WISDM\cite{wisdm_smartphone_and_smartwatch_activity_and_biometrics_dataset__507}  and OPPORTUNITY cover various sensor types, capturing diverse activities in daily life.
    \item \textbf{Single-node daily motion}: USC-HAD\cite{zhang2012usc}  serves as a standard benchmark, using a single right front hip sensor to evaluate basic human motion.
    \item \textbf{Elderly population}: HAR70 \cite{ustad2023validation}  captures the non-stationary motion rhythms of the elderly, used to validate the dynamic threshold modulation capability of TSLIF.
\end{itemize}

\begin{table*}[t]
\centering
\caption{Summary of selected IMU datasets, with window size \(T\) and step size adaptively configured according to the original sampling rate.}
\label{tab:dataset}
\resizebox{\linewidth}{!}{
\begin{tabular}{lccccccc}
\toprule
Dataset & Split & Subjects & Classes & Window Size(\(T\)) & Stride & Model input & Sensor Nodes(\(V\)) \\
\midrule
Daily-Sports & [0.70,0.15,0.15] & 8 & 19 & 125 & 50 & \(3 \times 5\) & 5 \\
HAR70 & [0.70,0.15,0.15] & 18 & 7 & 100 & 50 & \(3 \times 2\) & 2 \\
HuGaDB & [0.70,0.15,0.15] & 18 & 12 & 128 & 64 & \(6 \times 6\) & 6 \\
Opportunity & [0.70,0.15,0.15] & 4 & 4 & 60 & 30 & \(114 \times 1\) & 1 \\
PAMAP2 & [0.70,0.15,0.15] & 9 & 12 & 200 & 100 & \(6 \times 3\) & 3 \\
TNDA & [0.70,0.15,0.15] & 50 & 8 & 200 & 100 & \(3 \times 5\) & 6 \\
USC-HAD & [0.70,0.15,0.15] & 14 & 12 & 200 & 100 & \(3 \times 1\) & 1 \\
WISDM & [0.70,0.15,0.15] & 51 & 18 & 200 & 100 & \(6 \times 2\) & 2 \\
\bottomrule
\end{tabular}
}
\end{table*}

\subsubsection{Data Preprocessing}
All datasets adopt a frequency-adaptive sliding window strategy, with window size and stride dynamically adjusted based on the original sampling rate to ensure an approximately 2-second physical observation window. The Daily-Sports dataset retains its original 5-second pre-segmented snippets. Input signals are normalized to zero mean and unit variance, and converted to \(T \times B \times C \times V\) format to adapt to the model.
To avoid data leakage caused by window-level random splitting, we adopt a strict subject-independent hold-out protocol for all datasets except WISDM: each dataset is split by unique subject IDs; WISDM, due to its many fine hand movements and large inter-person variance, uses random splitting, with 70\% of subjects for training, 15\% for validation, and 15\% for testing. The test set contains only individuals unseen during training. All results are reported as accuracy scores to handle class imbalance.

\subsection{Overall Recognition Performance Comparison}
We compared TSFFormer with various state-of-the-art (SOTA) models, mainly in two categories:
\begin{itemize}
    \item Traditional Deep Neural Networks (ANN): DeepConvLSTM\cite{singh2020deep}.
    \item State-of-the-art Spiking Neural Networks (SNN): LMUformer\cite{liu2024lmuformer}, QKformer\cite{zhou2024qkformer}, Simple-SNN, SpikeGRU, SpikeRNN, SpikeTCN, Spikeformer\cite{zhou2022spikformer}, STATten\cite{lee2025spiking}, TSSNN \cite{yu2025ts}, SpikeDrivenTransformer\cite{yao2023spike}, and its improved version SpikeDrivenTransformerV2\cite{yao2024spike}.
\end{itemize}

Table~\ref{tab:performance} presents the recognition accuracy of all models across the eight benchmark datasets. TSFFormer establishes new SOTA performance in the vast majority of scenarios.

\begin{table*}[t]
\centering
\caption{Overall recognition performance (accuracy \%) on eight different IMU datasets. Our model is indicated in bold, and the ANN model is underlined.}
\label{tab:performance}
\resizebox{\linewidth}{!}{
\begin{tabular}{llcccccccc}
\toprule
Type & Model & Daily-sports & HAR70 & HuGaDB & Opportunity & PAMAP2 & TNDA & USC-HAD & WISDM \\
\midrule
ANN & \underline{DeepConvLSTM} & \underline{86.22} & \underline{90.60} & \underline{89.73} & \underline{94.10} & \underline{81.16} & \underline{92.28} & \underline{75.72} & \underline{71.23} \\
\midrule
\multirow{12}{*}{SNN} 
& LMUformer & 65.78 & 79.47 & 85.62 & 91.06 & 58.01 & 80.36 & 57.99 & 27.69 \\
& QKformer & 68.46 & 85.35 & 86.17 & 88.99 & 64.56 & 81.75 & 61.81 & 41.63 \\
& Simple-SNN & 52.54 & 77.22 & 81.47 & 88.75 & 57.40 & 71.84 & 51.26 & 36.62 \\
& SpikeGRU & 6.57 & 48.50 & 29.01 & 45.43 & 12.35 & 14.36 & 11.37 & 12.07 \\
& SpikeRNN & 64.07 & 86.91 & 89.55 & 90.45 & 64.60 & 79.78 & 53.93 & 20.07 \\
& SpikeTCN & 58.64 & 86.93 & 80.83 & 89.17 & 59.12 & 85.48 & 65.87 & 52.43 \\
& Spikeformer & 74.60 & 84.70 & 89.32 & 92.29 & 69.92 & 86.50 & 64.71 & 47.84 \\
& STATten & 79.03 & 83.26 & 87.63 & 89.32 & 64.05 & 86.25 & 68.82 & 42.25 \\
& TSSNN & 64.16 & 82.22 & 89.64 & 92.27 & 66.49 & 76.67 & 42.29 & 24.78 \\
& SpikeDrivenTransformer & 78.20 & 84.34 & 79.56 & 92.43 & 68.26 & 77.03 & 72.02 & 34.01 \\
& SpikeDrivenTransformerV2 & 79.82 & 86.21 & 89.73 & 88.62 & 71.55 & 93.12 & 67.00 & 49.79 \\
& \textbf{TSFFormer} & \textbf{87.89} & \textbf{90.54} & \textbf{90.42} & \textbf{93.79} & \textbf{83.33} & \textbf{98.53} & \textbf{78.65} & \textbf{86.35} \\
\bottomrule
\end{tabular}
}
\end{table*}

\textbf{Comprehensively surpassing existing SNN architectures:} TSFFormer achieves the best SNN performance on all 8 datasets. On the most challenging WISDM dataset, our model (86.35\%) achieves an absolute accuracy leap of 33.92\% compared to the second-best SpikeTCN (52.43\%); on the TNDA dataset, it improves by 5.41\% compared to SpikeDrivenTransformerV2 (93.12\%). This fully demonstrates the significant advantages of TSLIF dynamic neurons and Multi-Head Fourier attention in capturing long spatio-temporal dependencies and complex kinematic features.

\textbf{More remarkably, breaking the accuracy barrier between ANN and SNN:} TSFFormer not only surpasses all SNN baselines but also comprehensively defeats the best continuous DNN model, DeepConvLSTM, on datasets including Daily-sports (87.89\% vs. 86.22\%), PAMAP2 (83.33\% vs. 81.16\%), TNDA (98.53\% vs. 92.28\%), and WISDM (86.35\% vs. 71.23\%). This breaks the traditional stereotype that SNNs struggle to balance high accuracy and low power, proving that spiking architectures also possess powerful representation capabilities.This fully address RQ1 and demonstrates the superiority of TSFFormer.

\subsection{Training Efficiency and Convergence Analysis}
\begin{figure}[t]
\centering
\includegraphics[width=\linewidth]{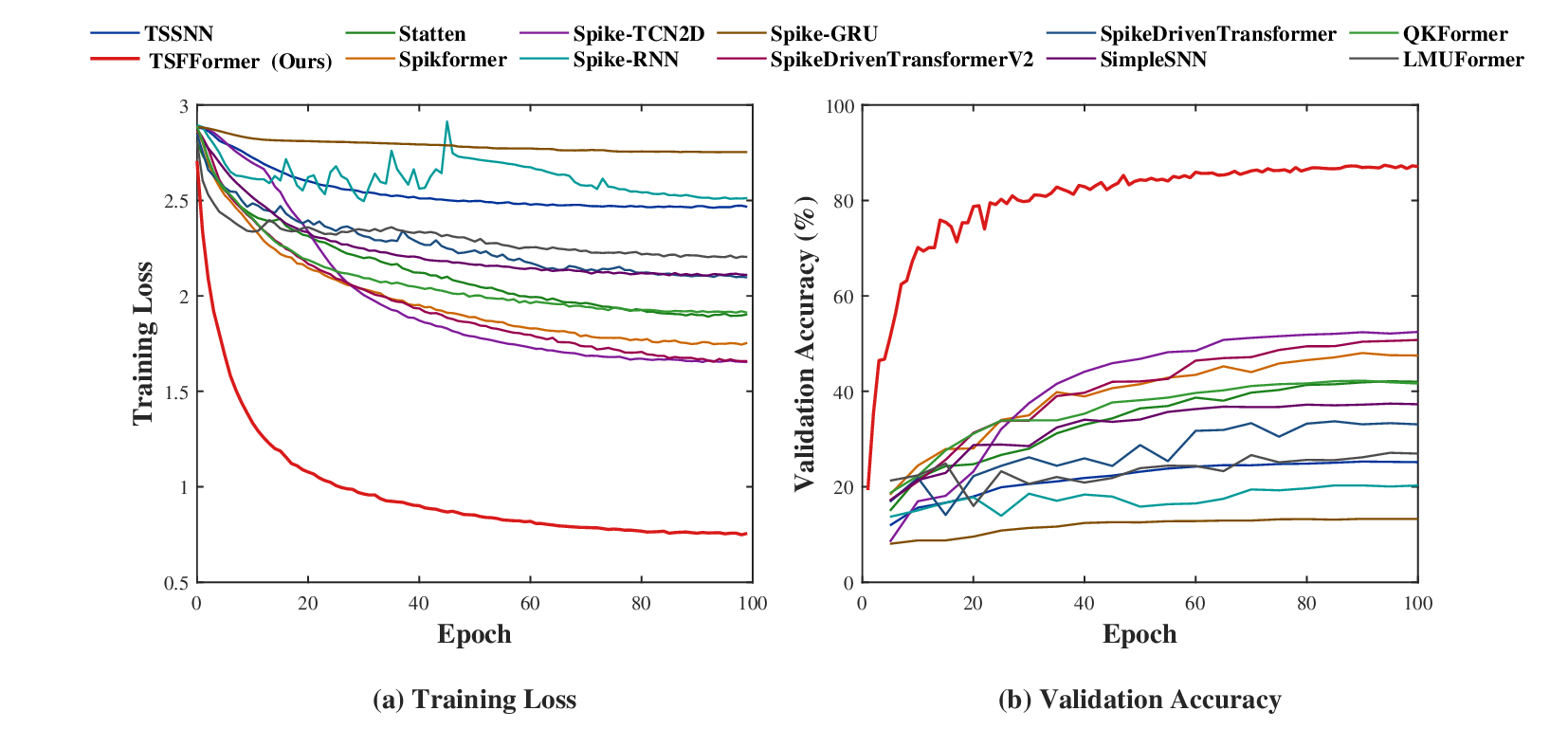}
\caption{Training loss curves and validation accuracy curves of the twelve models on the WISDM dataset.}
\label{fig:snns_dynamics}
\end{figure}

To investigate the training dynamics of TSFFormer and address RQ2, we compared the loss descent trajectories (Fig.~\ref{fig:snns_dynamics}a) and validation accuracy curves (Fig.~\ref{fig:snns_dynamics}b) of various models on the representative WISDM dataset.

\textbf{Our model exhibits extremely fast convergence and excellent gradient propagation:} As shown in Fig.~\ref{fig:snns_dynamics}(a), TSFFormer (Ours, red) experiences a cliff-like drop in loss values in the early training stage (\(<\)10 Epochs), rapidly converging to around 0.5. In contrast, traditional SNNs (e.g., Simple-SNN, Spike-RNN) and some Transformer architectures (e.g., SpikeDrivenTransformer) show extremely slow loss descent and struggle to converge to an ideal state. This indicates that the dual-compartment dynamics and threshold modulation mechanisms in TSLIF neurons effectively alleviate the common gradient degradation issue in deep SNN training via BPTT.

\textbf{Furthermore, our model exhibits high and stable validation accuracy:} In Fig.~\ref{fig:snns_dynamics}(b), the validation accuracy of TSFFormer rapidly climbs and stabilizes at a high level within a very short number of Epochs, without the severe oscillations or low accuracy traps common in early SNN models. This fast convergence property has extremely important engineering value for practical deployment and fine-tuning on edge devices.

\begin{figure}[t]
\centering
\includegraphics[width=\linewidth]{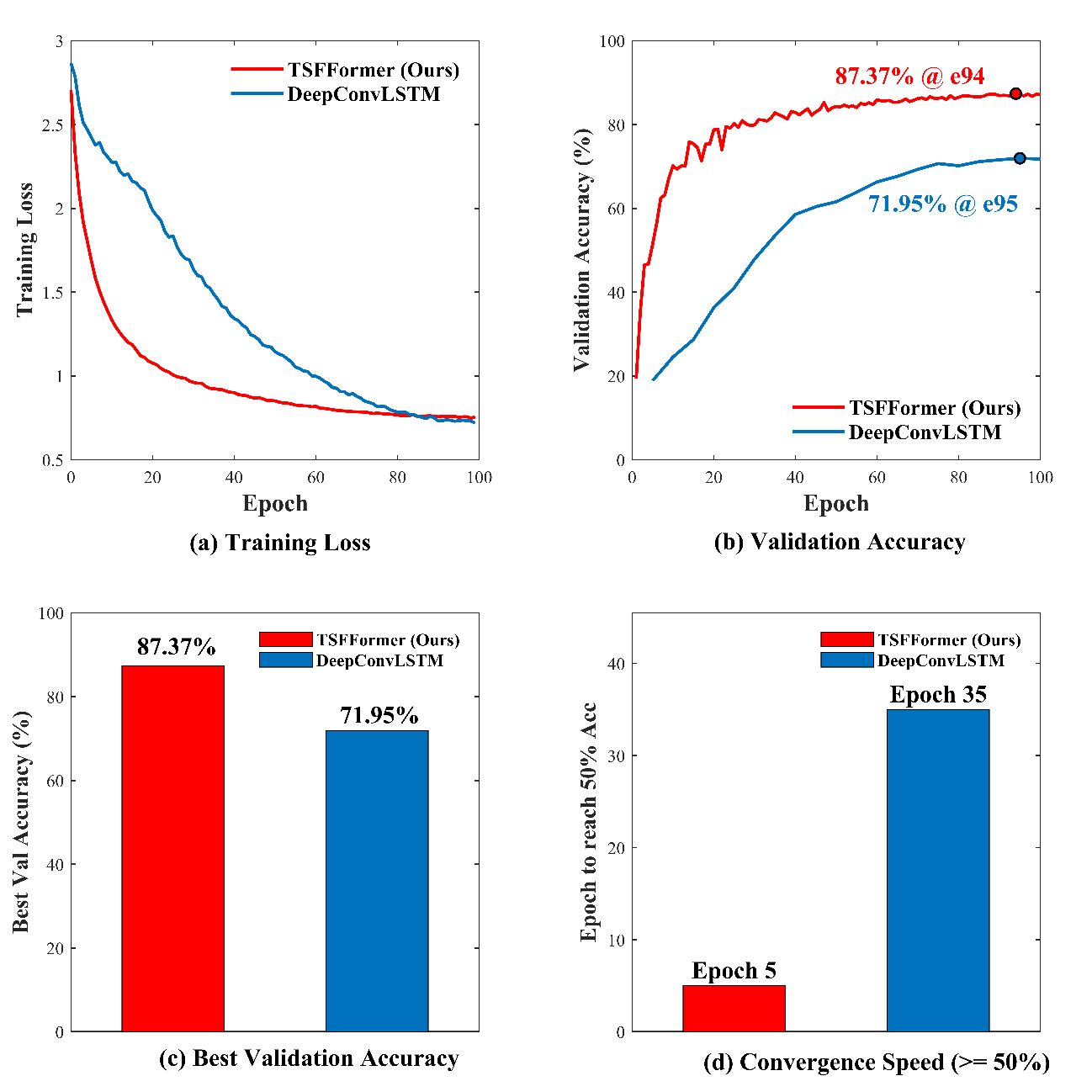}
\caption{Comparison of training dynamics and performance between TSFFormer and DeepConvLSTM on WISDM: (a) Training loss curve; (b) Accuracy trajectory of the validation set; (c) Comparison of the best validation accuracy; (d) Number of Epochs required to reach 50\% accuracy (convergence speed).}
\label{fig:dnn_dynamics}
\end{figure}

In addition, we not only compared various SNN baselines but also selected the highly representative continuous DNN, DeepConvLSTM, for a strict end-to-end training dynamics comparison on the same dataset. As shown in Fig.~\ref{fig:dnn_dynamics}, TSFFormer demonstrates crushing convergence advantages and a high representation ceiling.

As shown in the training loss curve in Fig.~\ref{fig:dnn_dynamics}(a) and the convergence speed comparison in Fig.~\ref{fig:dnn_dynamics}(d), TSFFormer loss undergoes a cliff-like drop in the early training stage, requiring only 5 Epochs to break through the 50\% validation accuracy threshold. In contrast, DeepConvLSTM takes a lengthy 35 Epochs to reach the same level, meaning r the early convergence speed of TSFFormer is 7 times that of the traditional DNN. This is not only due to the efficient feature routing brought by frequency-domain global mixing but also proves that the dynamic dual-compartment mechanism of TSLIF neurons effectively mitigates gradient vanishing in deep networks during temporal propagation.

From the validation accuracy trajectory in Fig.~\ref{fig:dnn_dynamics}(b),  the accuracy curve of TSFFormer rapidly climbs and stabilizes at a high level without the severe oscillations common in traditional DNNs. Ultimately, as shown in Fig.~\ref{fig:dnn_dynamics}(c), TSFFormer achieves a best validation accuracy of 87.37\%, while DeepConvLSTM remains at 71.95\%, giving TSFFormer an absolute accuracy improvement of 15.42\%.This overturns the long-held assumption that SNNs cannot match the accuracy of traditional DNNs, underscoring the immense viability of purely event-driven paradigms for wearable sensing.

\subsection{Ablation Study}
To rigorously evaluate the effectiveness of the core architecture of TSFFormer and address RQ3, we constructed four progressive variants for strict ablation studies.
\begin{itemize}
    \item TSFFormer (Ours):The complete proposed model;
    \item w/o MultiHeadFourier: replacing the frequency-domain mixing mechanism with standard spatial self-attention;
    \item w/o TSLIFNode: replace the dual-compartment dynamic neuron with a fixed-threshold standard LIF neuron;
    \item w/o Both (pure baseline SNN): removing both the frequency-domain mechanism and the dynamic neuron.
\end{itemize}
We comprehensively evaluated them on two dimensions: recognition accuracy and training dynamics (convergence speed and computational overhead), with results shown in Tables~\ref{tab:ablation_acc} and \ref{tab:ablation_eff}.

\begin{table*}[t]
\centering
\caption{Ablation experiment results of core components on eight benchmark datasets (accuracy \%).}
\label{tab:ablation_acc}
\resizebox{\linewidth}{!}{
\begin{tabular}{lcccccccc}
\toprule
Model & Daily-Sports & HAR70 & HuGaDB & Opportunity & PAMAP2 & TNDA & USC-HAD & WISDM \\
\midrule
TSFFormer & 87.89 & 90.54 & 90.42 & 93.79 & 83.33 & 98.25 & 78.65 & 86.35 \\
w/o MultiHeadFourier & 86.09 & 88.95 & 88.50 & 92.25 & 80.85 & 96.89 & 78.94 & 81.18 \\
w/o TSLIFNode & 80.13 & 88.03 & 90.14 & 91.85 & 79.76 & 93.12 & 71.39 & 63.85 \\
w/o Both & 64.03 & 83.94 & 86.86 & 87.96 & 61.48 & 77.36 & 54.14 & 26.07 \\
\bottomrule
\end{tabular}
}
\end{table*}

\begin{table*}[t]
\centering
\caption{Comparison of training efficiency, convergence, and system resource consumption of the model on the TNDA dataset.}
\label{tab:ablation_eff}
\resizebox{\linewidth}{!}{
\begin{tabular}{lccccc}
\toprule
Model configuration & Epochs to 95\% & Mean epoch time (s) & Peak GPU memory (MB) & Best validation (\%) & Time to best validation (h) \\
\midrule
TSFFormer & 22 & 188.5 & 10905 & 96.60 & \(\sim\)2.6 \\
w/o MultiHeadFourier & 39 & 567.2 & 20713 & 95.45 & \(\sim\)12.4 \\
w/o TSLIFNode & Not Achieved & 112.2 & 7968 & 92.71 & \(\sim\)1.34 \\
w/o Both & Not Achieved & 294.91 & 10334 & 78.19 & \(\sim\)7.13 \\
\bottomrule
\end{tabular}
}
\end{table*}

\subsubsection{ Effectiveness of TSLIFNode}
Removing TSLIFNode caused the most severe performance degradation. As shown in Table~\ref{tab:ablation_acc}, on the highly dynamic WISDM dataset, accuracy plummeted from 86.35\% to 63.85\% (a drop of over 20\%); and from Table~\ref{tab:ablation_eff}, the w/o TSLIFNode model even failed to reach the 95\% accuracy line (peaking at only 92.71\%).

This phenomenon profoundly reveals that the fixed threshold of standard LIF neurons is completely incapable of handling non-stationary, high-burst human motion rhythms, easily leading to ``spike explosions'' or loss of resting features. Although TSLIF introduces dual-compartment dynamics that slightly increase per-epoch time to 188.5s, it brings a massive improvement in convergence speed (reaching 95\% in just 22 epochs), making it the core engine for high robustness.

\subsubsection{Impact of MultiHeadFourier}
Removing MultiHeadFourier also leads to general accuracy drops (e.g., HuGaDB dropping by 3.88\%, PAMAP2 by 2.48\%). More critically, Table~\ref{tab:ablation_eff} shows that traditional spatial dot-product attention incurs catastrophic computational overhead: its per-epoch training time soars to 567.2s, and peak memory usage reaches 20,713 MB. In contrast, MultiHeadFourier uses the Fast Fourier Transform (\(O(V\log V)\) complexity) to perform element-wise multiplication in the frequency domain, increasing training speed by 3 times, saving nearly 50\% of memory, and requiring only 22 epochs to break through 95\% accuracy. This fully proves that the frequency-domain mechanism, while capturing periodic topological dependencies between multiple sensors, perfectly aligns with the ``green and low-power'' design philosophy. (Note: On the single-node USC-HAD dataset, removing FFT actually yields a slight increase of 0.29\%. This is mainly because a single node lacks complex spatial topology, and frequency-domain global mixing may introduce slight redundancy; while removing TSLIF on this dataset causes a plunge of 7.26\%, once again confirming the fundamental role of temporal dynamic modulation.)

\subsubsection{ Pure Baseline}
The w/o Both pure baseline model (using standard self-attention and fixed-threshold LIF) shows complete collapse across all metrics. In Table~\ref{tab:ablation_acc}, it achieves catastrophic accuracies of only 61.48\% and 26.07\% on PAMAP2 and WISDM, respectively; in Table~\ref{tab:ablation_eff}, it not only fails to reach 95\% accuracy (peaking at only 78.19\%), but due to gradient degradation and the superposition of attention matrices, its per-epoch time (294.9s) is even slower than the full model.

This controlled variable experiment forms a perfect logical loop: TSLIFNode provides fine-grained spike sparsification and adaptive modulation in the temporal dimension, while MultiHeadFourier provides a global spatial receptive field in the frequency domain. Neither can be dispensed with; only by deeply combining the two can we simultaneously achieve extremely fast convergence (\(\sim\)2.6 hours), ultra-low memory usage (\(\sim\)10.9 GB), and the highest classification accuracy (96.60\%) under the strict spiking (pure AC operation) paradigm.
\section{Conclusion}
We propose an end-to-end trained purely spiking architecture, TSFFormer, which applies frequency-domain global mixing \cite{gan2026caracal} and dynamic neural modulation\cite{feng2025ts} to effectively balance computational cost and recognition performance. The performance of TSFFormer surpasses state-of-the-art baselines in the IMU-HAR domain (even outperforming dense DNNs), while featuring higher convergence speed (3\(\times\) faster) and lower memory usage (\(\sim\)50\% savings). Furthermore, we incorporate MultiHeadFourier attention and dual-compartment TSLIFNode for spatio-temporal feature extraction, resulting in a highly robust event-driven model that exhibits excellent accuracy on eight benchmark datasets while retaining the ultra-low power consumption of pure AC operations. Our frequency-spiking hybrid structure is highly promising for deployment on resource-constrained wearables and event-driven neuromorphic processors. We believe that the combination of frequency-domain attention and dynamic spiking dynamics holds promise for migration to broader multimodal time-series tasks, and the proposed TSFFormer is of significant value for promoting the development of Green HAR.
{
    \small
    \bibliographystyle{ieeenat_fullname}
    \bibliography{main}
}


\end{document}